\documentclass{article}
\usepackage{iclr2027_conference,times}
\usepackage[T1]{fontenc}
\usepackage[utf8]{inputenc}
\usepackage{graphicx,booktabs,array,tabularx,longtable,multirow,float,caption}
\usepackage{amsmath,amssymb}
\usepackage{enumitem}
\usepackage{xcolor}
\usepackage{listings}
\usepackage[hidelinks]{hyperref}
\usepackage{url}
\makeatletter
\expandafter\def\csname bibstyle@iclr2027_conference\endcsname{\bibpunct{(}{)}{;}{a}{,}{;}}
\def\paragraph{\@startsection{paragraph}{4}{\z@}{0.1pt}{-1em}{\normalsize\bf}}
\makeatother

\graphicspath{{figures/redrawn/}{figures/}}
\definecolor{accent}{HTML}{287D8E}
\definecolor{linkblue}{HTML}{1F4E9E}
\definecolor{linkgray}{HTML}{6B7280}
\usepackage{tikz}
\newcommand{\webicon}{\raisebox{-0.12em}{\begin{tikzpicture}[x=1em,y=1em,line width=0.07em,draw=linkblue]
  \draw (0,0) circle (0.45);
  \draw (0,0) ellipse (0.2 and 0.45);
  \draw (-0.45,0) -- (0.45,0);
  \draw (-0.39,0.22) -- (0.39,0.22); \draw (-0.39,-0.22) -- (0.39,-0.22);
\end{tikzpicture}}}
\newcommand{\hficon}{\IfFileExists{figures/hf-logo.png}{\raisebox{-0.22em}{\includegraphics[height=1.15em]{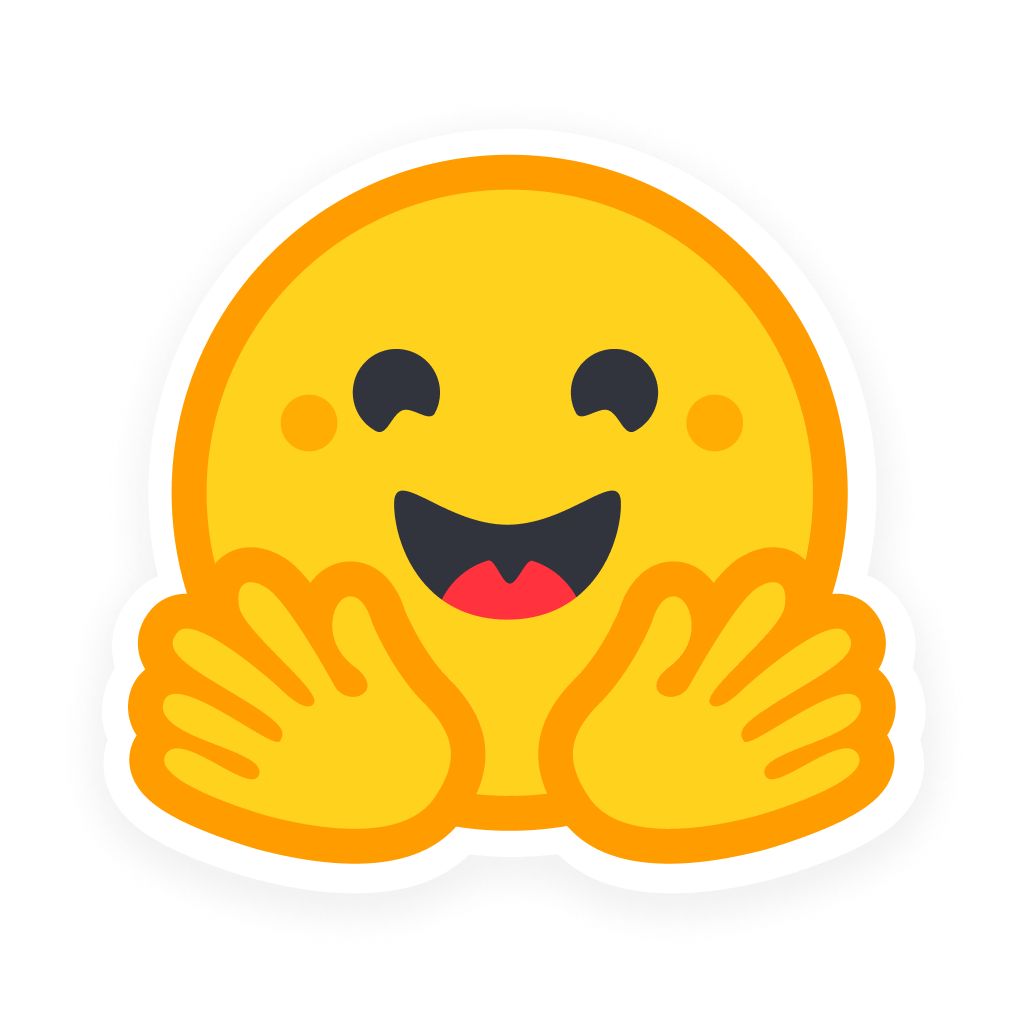}}}{}}
\newcommand{\resourcelink}[4]{%
  \href{#3}{\begin{tabular}{@{}c@{}}{\color{linkblue}#1\hspace{0.3em}\textsf{\textbf{#2}}}\\[-1pt]{\scriptsize\ttfamily\color{linkgray}#4}\end{tabular}}}
\newcommand{\nextme}{NextMe-800}
\newcommand{\nextact}{NextAct}

\title{NextMe-800: Anticipating Personal Behavior from Months of Egocentric Video}
\author{Zhaoxu Meng$^{1}$ \quad Yiming Sun$^{1}$ \quad Mingyuan Gao$^{1}$ \quad Jiachang Zhang$^{1}$ \\
\textbf{Zhuhan Dai$^{1}$ \quad Yipeng Du$^{1}$ \quad Zheng Lian$^{2}$ \quad Jian-Qiao Zhu$^{1}$} \\
$^{1}$The University of Hong Kong \qquad $^{2}$Tongji University \\
}
\iclrfinalcopy
\begin{document}
\raggedbottom
\maketitle
\lhead{Preprint. Under review.}
\begin{abstract}
We often plan ambitiously yet act habitually and wonder, in retrospect, whether we would have planned differently had we known what we would actually do. Hindsight offers a valuable perspective on past decisions, although we often wish we could have simulated hindsight at the moment of choosing. If a system could generate plausible trajectories from one's personal history, such previews might help people formulate more realistic plans and make better informed decisions. We introduce \nextme{}, an approximately 800-hour first-person dataset from one volunteer over 126 days with 1\,Hz images, gaze, and audio, captioned at five hierarchical abstraction levels from atomic actions to major activities. We formulate personalized action anticipation as open-vocabulary $K$-step sequence prediction and construct \nextact{}, a 1{,}500-point benchmark combining \nextme{} with the multi-person EgoLife dataset~\citep{egolife2025}. Using an embedding-based soft edit distance as the metric, we evaluate how well different models can anticipate personal behavior across abstraction levels and prediction horizons. \nextme{} and \nextact{} provide a months-long resource and evaluation framework for studying how far ahead personal behavior can be anticipated from egocentric observation.
\end{abstract}
\vspace{-2pt}\begin{center}
\resourcelink{\hficon}{Dataset \& Benchmark}{https://huggingface.co/datasets/mmm8383/NextMe-800}{huggingface.co/datasets/mmm8383/NextMe-800}%
\hspace{2.2em}%
\resourcelink{\webicon}{Project Page}{https://kkkkawayi.github.io/nextme-800/}{kkkkawayi.github.io/nextme-800}
\end{center}\vspace{-4pt}
\begin{figure}[H]
\centering
\includegraphics[width=0.9\linewidth]{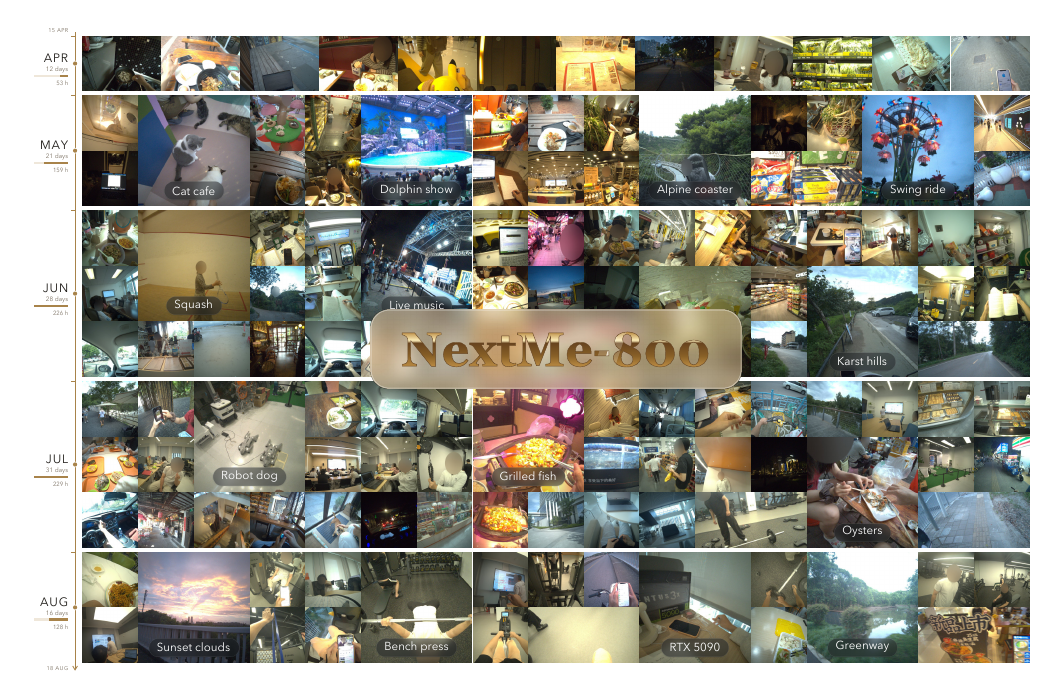}
\caption{\textbf{Overview of the \nextme{} dataset.} Nearly 800 hours of in-the-wild egocentric video provide a comprehensive record of four months of a single individual's life, capturing a rich diversity of activities spanning both habitual routines and long-tail activities. 
}
\label{fig:nextme-overview}
\end{figure}
\newpage
\section{Introduction}

\begin{quote}
\textit{``Are these the shadows of the things that Will be,
or are they shadows of things that May be, only?''}
\par\raggedleft
Charles Dickens, \textit{A Christmas Carol}
\citeyearpar[Stave~IV]{dickens1843christmas}
\end{quote}

We often make ambitious plans but struggle to anticipate how our future selves will actually behave. Initial enthusiasm for learning the piano or pursuing a research project may give way to familiar routines. Hindsight helps us re-evaluate these decisions, but we wish we could simulate that perspective at the moment of choosing. If a system could preview plausible behavioral trajectories generated from one's personal history, it might help people formulate more realistic plans. For example, it could tell someone in the morning, \emph{``Based on your past patterns, you will likely start work this afternoon, get distracted by your phone, and only resume in the evening.''} Indeed, vivid encounters with future selves have been shown to reshape financial decisions~\citep{hershfield2011}, and recent AI-generated digital twins that simulate life paths significantly strengthen future self-continuity and expand human choice~\citep{futureyou2025}. Making such naturalistic previews reliable requires sustained, detailed observations of personal routines and ways to assess whether predictions match what actually happens, from immediate actions to broader activities. This raises the question: \emph{what data and evaluation framework are needed to support reliable, open-vocabulary prediction of a person's future behavior from months of egocentric experience?}

Continuous egocentric recording is a natural medium for addressing these questions: it captures what a person actually does, hour by hour, with rich contextual detail. This modality has attracted growing research interest for augmenting personal memory through always-on assistants that remind the wearer of forgotten items or past events~\citep{ego4d2022}, and for providing demonstrations for robot learning~\citep{kareer2025egomimic}. We argue that anticipating a person's future behavior, and thereby helping them reflect on their likely choices before committing, is another promising application. Dense observation helps capture brief actions, interruptions, and transitions that sparse snapshots may miss. Following the same person over months allows us to observe how these behaviors recur across contexts, including less frequent activities and changes in routine. However, existing resources address different aspects of this problem, providing broad egocentric diversity and an action-anticipation benchmark \citep{ego4d2022}, rich multi-person, multi-day observations \citep{egolife2025,castle2025}, personal memory evaluation across weeks and months \citep{egomonth2026}, and personalized next-action prediction from digital interactions \citep{longnap2026}.

To support behavioral forecasting, we need resources that combine dense sampling and extended temporal coverage with sufficient visual resolution to capture both physical and digital activities. Recording gaps must also be explicitly documented, and behavior represented at multiple levels of abstraction, from immediate actions to broader activities. Such temporal depth within one individual complements diversity across individuals by enabling repeated observations of how the same person behaves across situations and over time.

We introduce \nextme{}, an approximately 800-hour first-person collection across 126 days. The record covers study, work, meals, travel, exercise, social interaction, and leisure. High-resolution RGB supports descriptions of screen-mediated actions, while gaze and audio provide complementary evidence. A de-identification and captioning pipeline converts the recordings into timestamped descriptions, and a reference-preserving hierarchy represents immediate actions through major activities. In addition, \nextact{} combines fixed evaluation points from this collection and EgoLife \citep{egolife2025}. Its prediction target is the next $K$ \emph{recorded} events at a specified abstraction level, conditioned on the target recording windows; unobserved gaps are not treated as known behavior.

Evaluation is central yet more challenging for open-vocabulary, naturalistic events: exact string matching rejects valid paraphrases, whereas broad semantic similarity can reward the correct scene but the wrong action. We therefore validate automatic scorers against human rankings before using semantic similarity as the substitution cost in edit distance.

Our contributions are threefold. First, we provide a months-long egocentric resource with an explicit account of temporal coverage, annotation, and privacy preservation. Second, we formulate multi-resolution, open-vocabulary anticipation and document an embedding-based sequence metric, including human-agreement and score-distribution diagnostics. Third, we benchmark repetition, frontier language models, and a retrieval-augmented agent on \nextact{}, and analyze how predictability varies with abstraction level and prediction horizon.

\section{Related Work}
\label{sec:related}

\paragraph{Long-horizon action anticipation.}
Action anticipation has progressed from next-action classification to forecasting future activities and their timing~\citep{furnari2020rolling,abufarha2018when}. A common formulation predicts sequences of verb--noun pairs from a fixed taxonomy, evaluated using edit distance~\citep{ego4d2022}. Recent approaches combine visual descriptions with language modeling and goal inference while retaining this closed-vocabulary formulation~\citep{palm2024,antgpt2024}. Personalized, open-vocabulary prediction has also been explored using one month of phone-screen actions from 20 users, although these observations are limited to digital interactions~\citep{longnap2026}. Our dataset provides a much longer record of everyday physical and digital activities. Its five abstraction level of behavioral hierarchy supports open-vocabulary anticipation and enables studying whether weeks or months of personal history improve predictions beyond the immediate context.

\paragraph{Longitudinal egocentric datasets.}
Longitudinal egocentric recordings range from nine months of sparse outdoor frames \citep{krishnacam2016} to hundreds of hours of child-perspective video \citep{saycam2021} for developmental research. Lifelog retrieval benchmarks typically emphasize still images sampled every 30~seconds~\citep{gurrin2021experiments,lsc2026}. Dense multi-person recordings cover up to one week~\citep{egolife2025,castle2025}, while personal memory benchmarks extend across months with sparser sampling~\citep{egomonth2026}. Complementary benchmarks support goal-based understanding of procedural activities and planning from egocentric video~\citep{goalstep2023,egoplan22024}. Our dataset combines temporal depth with 1 Hz images, gaze, and audio. The resulting abstraction levels span atomic actions to major activities without a prescribed taxonomy, supporting open-vocabulary prediction of how everyday behavior actually unfolds.

\paragraph{Ultralong video understanding and behavior simulation.}
Structured multimodal memory supports retrieval and question answering over recordings spanning hours to weeks~\citep{worldmm2025,magicvideo2026}. Extending these systems to behavior forecasting requires restricting retrieval to evidence available before the prediction point. LLMs coupled with memory and environment modules can simulate interactive behavior and emergent social dynamics~\citep{generativeagents2023,concordia2023}. However, plausible simulations do not by themselves demonstrate accurate prediction of an individual's future actions. Our longitudinal dataset enables direct evaluation against observed future behavior, using an embedding-based metric normalized against random baselines.

\section{Dataset Construction: What 800 Hours of Life in Egocentric Video Looks Like}
\label{sec:data}

\paragraph{Collection and design choices.}
To capture the full detail of one person's daily behavior as continuously as possible, \nextme{} records one volunteer wearing Project Aria glasses~\citep{aria2023} across 126 calendar days, producing 795.5 hours of first-person video, an average of 6.3 hours per day. RGB is sampled at 1\,Hz at $2880\times2880$ resolution, accompanied by continuous gaze and audio streams, totaling 8.5\,TB .

There are some considerations for selecting such parameters: 1\,Hz sampling also balances behavioral coverage against cost: most vision-language models downsample to roughly 1\,fps for video understanding, and higher frame rates bring diminishing informational returns for action-level description.
$2880\times2880$ resolution ensures that text on phone and laptop screens remains legible, preserving the digital half of modern daily behavior.
Gaze tracking provides continuous attention information that disambiguates what the wearer is focusing on when multiple objects are in view.
Recording spans from shortly after waking until bedtime, with intermittent pauses for charging or incompatible activities; each recorded day therefore comprises multiple disjoint sessions (4.9 on average across 528 total sessions), and 18 of the 126 calendar days contain no recording at all. Multi-step predictions routinely bridge both intra-day gaps between sessions and missing days.

\paragraph{What activities does the collection cover?}
The recording broadly covers both habitual routines and long-tail events (see Figure \ref{fig:nextme-overview}). Recurring activities include academic research, coding, meals, dating, sports, and guitar practice. Long-tail sequences range from boarding flights, watching the sunset from a ferry, and exploring karst landscapes to playing squash, observing dolphins and beluga whales, and interacting with robotic dogs. Together, these span diverse physical environments, social contexts, cognitive tasks, and personal interests, revealing how one individual's routines and deviations evolve over months.

\begin{figure}[t]
\centering
\begin{minipage}[t]{0.56\linewidth}
\centering{\small (a) Daily recording: 795.5\,h over 126 days}\par\vspace{3pt}
\includegraphics[width=\linewidth]{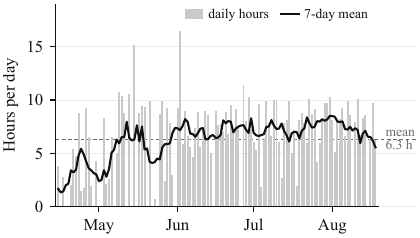}
\end{minipage}\hfill
\begin{minipage}[t]{0.41\linewidth}
\centering{\small (b) Dataset-wide compression}\par\vspace{18pt}
{\footnotesize\setlength{\tabcolsep}{4pt}\renewcommand{\arraystretch}{1.15}
\begin{tabular}{@{}lrrrr@{}}
\toprule
 & & Mean & \multicolumn{2}{c}{Ratio} \\
\cmidrule(l){4-5}
Level & Events & dur. & Adj. & Cum. \\
\midrule
L1 & 529{,}431 & ${\sim}$5\,s & --- & --- \\
L2 & 48{,}093 & ${\sim}$1\,min & 11$\times$ & 11$\times$ \\
L3 & 6{,}367 & ${\sim}$7.5\,min & 7.6$\times$ & 83$\times$ \\
L4 & 1{,}988 & ${\sim}$24\,min & 3.2$\times$ & 266$\times$ \\
L5 & 1{,}236 & ${\sim}$39\,min & 1.6$\times$ & 428$\times$ \\
\bottomrule
\end{tabular}}
\end{minipage}
\caption{Temporal coverage and annotation scale. (a) Recorded hours per calendar day (gray bars), their 7-day moving average (solid black line), and the overall daily mean (dashed gray line); days without recordings are kept. (b) Event counts, mean event duration, and adjacent and cumulative compression ratios at each abstraction level.}
\label{fig:collection}
\end{figure}

\paragraph{Why this temporal scale?}
Table~\ref{tab:datasets} compares longitudinal recording statistics. Ego4D provides broad diversity across 931 participants but averages only 3.9 hours per person, precluding longitudinal study of any individual. EgoLife achieves dense waking-hour coverage (39.6\%) comparable to ours but spans only 7~days~\citep{egolife2025}. KrishnaCam follows one person for nine months yet captures only sparse outdoor frames. LongNAP records 28~days of screen interactions for 20 users but is limited to the digital domain. NTCIR-18/LSC achieves the longest span (547~days) and highest nominal coverage but samples one still image every 30~seconds without gaze or audio. \nextme{} combines months-long temporal depth with 1\,Hz video, gaze, and audio for a single individual, achieving 39.5\% awake-time coverage---recorded hours divided by waking hours assuming 16\,h awake per day---across 126 calendar days (see Figure~\ref{fig:collection}).

\begin{table}[t]
\centering\small
\caption{Longitudinal recording statistics, sorted by average per-person recording time. Awake coverage assumes 16 waking hours per day. $\dagger$\,=\,image-based estimate.}
\label{tab:datasets}
\setlength{\tabcolsep}{3pt}
\begin{tabular}{@{}l r r r r r l@{}}
\toprule
\multirow{2}{*}{Dataset} & {Avg.\ rec.} & {Avg.} & {Awake} & \multirow{2}{*}{People} & {Total} & \multirow{2}{*}{Notes} \\
 & {time (h)} & {span} & {cov.\ (\%)} & & {time (h)} & \\
\midrule
Ego4D \citep{ego4d2022} & 3.94 & -- & -- & 931 & 3{,}670 & POV video \\
EgoMonth \citep{egomonth2026} & 15.06 & 66.5\,d & 1.42 & 20 & 301.2 & POV video \\
EgoLife \citep{egolife2025} & 44.33 & 7\,d & 39.58 & 6 & 266 & POV video \\
KrishnaCam \citep{krishnacam2016} & 70.20 & ${\approx}$9\,mo & ${\approx}$1.63 & 1 & 70.2 & Outdoor POV \\
LongNAP \citep{longnap2026} & 91.85 & 28\,d & 20.50 & 20 & 1{,}837 & Computer screen rec. \\
SAYCam \citep{saycam2021} & $>$166.7 & ${\approx}$21.3\,mo & ${\gtrsim}$1.6 & 3 & $>$500 & Child POV \\
NTCIR-18/LSC \citep{ntcir18lifelog} & ${\approx}$6{,}044$^\dagger$ & 547\,d & ${\approx}$69.1$^\dagger$ & 1 & ${\approx}$6{,}044$^\dagger$ & POV stills; 0.033\,Hz \\
\midrule
\textbf{\nextme{} (ours)} & \textbf{795.50} & \textbf{126\,d} & \textbf{39.46} & \textbf{1} & \textbf{795.5} & \textbf{POV 1\,Hz; gaze; 2880$^2$} \\
\bottomrule
\end{tabular}
\end{table}

\paragraph{Privacy-aware processing.}
The de-identification pipeline prioritizes preserving behavioral information while removing direct identifiers. Detected faces are masked via CenterFace~\citep{centerface2020}; on-screen sensitive text (credentials, phone numbers, addresses) is located by PP-OCR~\citep{ppocrv6} and classified by the OpenAI Privacy Filter model~\citep{openaiprivacyfilter2026}, with only identified sensitive spans masked---entertainment content, work interfaces, and message text that describe behavior are retained. Speech is transcribed to identify and mute sensitive segments, and voice characteristics are transformed to reduce speaker identifiability. Together, these measures provide comprehensive protection across visual, textual, and audio modalities, although subtle contextual cues may persist in a months-long record.
\section{From Raw Recordings to Behavioral Data}
\label{sec:representation}
\vspace{-1pt}
\begin{figure}[H]
\centering\includegraphics[width=0.9\linewidth]{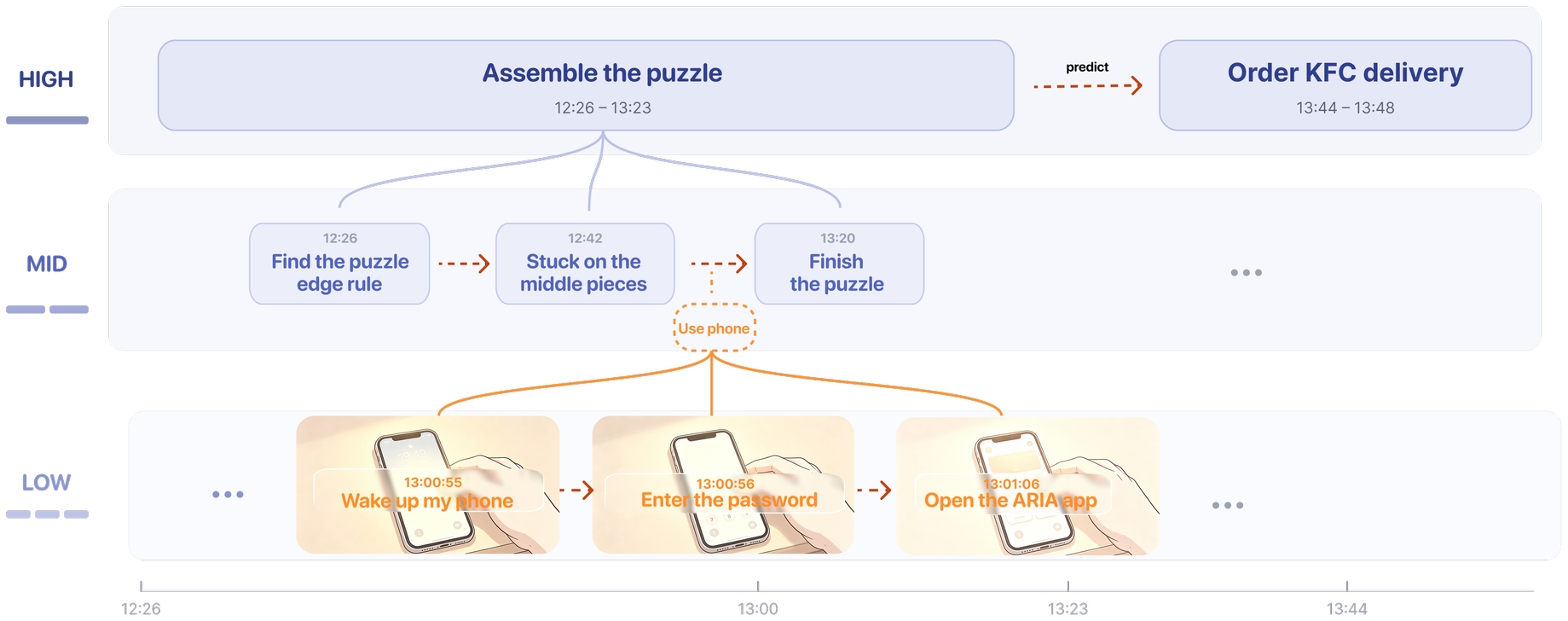}
\caption{Simplified illustration of the hierarchical caption tree, showing three of the five abstraction levels. At the lowest level, every few-second interaction is preserved (e.g., waking the phone, entering a password, opening an app). The mid level absorbs such incidental phone use and tracks subtask progression within the puzzle session. The high level collapses the entire hour-long session into a single event and the prediction target becomes the next major activity.}
\label{fig:hierarchy}
\end{figure}

This section describes the captioning pipeline that produces atomic action descriptions and the hierarchical abstraction that organizes them into five semantic levels.

\paragraph{Dense captioning.}
The caption pipeline processes 30-second windows. Each window consists of 1\,Hz sampled frames with the gaze trajectory rendered on each image, OCR-extracted screen text, and an aligned speech transcript. Because on-screen text is often small at the native $2880\times2880$ resolution, a cropped region around the gaze fixation point is additionally provided to the model. We use Gemini-family models~\citep{gemini35flash2026,gemini37flash2026} as the captioning backbone. The model returns first-person action descriptions that constitute L1 events in the abstraction hierarchy, and environment descriptions that establish the scene and subsequently record state changes. To assess caption quality, four annotators independently audit 613 caption segments, with injected erroneous captions as quality controls. After excluding one low-performing annotator, 88.25\% of source segments receive no error flag; the remaining errors mostly concern minor perceptual details, such as left--right hand confusion, rather than the core action (see Appendix~\ref{app:caption}).

\paragraph{Hierarchical caption tree.}
Raw L1 captions are inherently noisy: a three-hour study session produces thousands of atomic events including scratching one's nose, glancing at a phone, or adjusting the air conditioner. Events at lower abstraction levels preserve more information but are noisier and less predictable, whereas events at higher levels cover the same time span with fewer autoregressive steps, reducing compounding error at the expense of detail. No single level is optimal, so the abstraction hierarchy allows switching granularity to balance noise, information content, and compounding error as needed.

Following the recursive captioning paradigm~\citep{videorecap2024}, we organize captions into a five-level hierarchical tree (L1--L5) with adjacent-level compression ratios ranging from 1.6$\times$ to 11$\times$. Starting from L1 atomic actions, an LLM recursively abstracts each level into the next by grouping events that share a common goal using the same prompt structure at every transition (see Appendix~\ref{app:prompts}). Rather than feature-based clustering, the model decides merge boundaries based on semantic goal shifts; we impose no fixed time-scale range or minimum event duration, as constraining durations causes models to rigidly segment toward the prescribed range rather than following natural event boundaries. Each event at every level carries a single action description expressing one unified objective instead of a concatenation of child actions. ``Organize purchased groceries,'' for example, subsumes unpacking, sorting, and storing as a single parent. The resulting hierarchy provides a multi-resolution view of naturalistic behavior, enabling models to zoom between atomic detail and high-level routine structure depending on the prediction horizon (Figure~\ref{fig:hierarchy}; Figure~\ref{fig:collection}b).

\section{Evaluating Open-Vocabulary Action Sequences}
\label{sec:metrics}

How similar are a ground-truth action ``I grab a bottle of cola from the fridge'' and a model prediction ``I pick up a glass of water from the table''? We evaluate a variety of scoring methods, verify each against human agreement, and select embedding cosine similarity as the atomic action metric. For multi-step evaluation, following the edit-distance formulation in long-horizon action anticipation~\citep{ego4d2022}, we extend this to a soft edit distance that aligns two variable-length action sequences while tolerating insertions, deletions, and partial mismatches.

\paragraph{Atomic action similarity.}
A good evaluation metric should both agree with human judgment on which prediction is closer to the ground truth, and spread scores enough to discriminate between answers of different quality rather than piling them near 0 or 1. We formalize these as two diagnostics. \emph{Pairwise accuracy} measures ranking agreement: four annotators independently rank candidates by similarity to the ground truth, ties permitted; for each strict human preference between two candidates, a metric is correct if it assigns the preferred one a strictly higher score. \emph{Normalized score entropy} $H_{10}$ measures discrimination: all scores are binned into ten equal-width intervals on $[0,1]$ and the Shannon entropy of bin occupancy is normalized by $\log 10$~\citep{shannon1948}; low entropy indicates a metric that compresses most scores into a narrow band and cannot supply graded substitution costs for sequence alignment.

To construct a validation set that broadly covers the similarity spectrum, we collect 350 time points, each pairing a ground-truth action with four candidates. Candidates are drawn from two sources: real predictions from multiple models (GPT-5.4, DeepSeek-V4-Flash, Qwen3.5-35B) across diverse test points spanning different participants, granularity levels, and dates; and controlled rewrites of the ground truth at varying semantic distances (i.e., paraphrases that preserve meaning and partial rewrites that alter the verb or object). Real predictions alone cluster at low similarity, so the rewrites fill the upper range to ensure the full score distribution is represented.

\begin{figure}[t]
\begin{minipage}[b]{0.52\linewidth}
\centering\footnotesize
\setlength{\tabcolsep}{4pt}
\begin{tabular}{@{}lrrr@{}}
\toprule
Method & Pairwise $\uparrow$ & Top-1 $\uparrow$ & $H_{10}$ $\uparrow$ \\
\midrule
Qwen3-Reranker-8B & 0.9417 & 0.9636 & 0.6429 \\
Qwen3-Embedding-8B & 0.9290 & 0.9407 & 0.9724 \\
\cmidrule(lr){1-4}
Gemini-3.7-Flash & 0.9097 & 0.9571 & 0.9407 \\
GPT-5.6 & 0.8968 & 0.9489 & 0.9113 \\
Claude-Haiku-4.5 & 0.8952 & 0.9435 & 0.9612 \\
Qwen3.5-Flash & 0.8794 & 0.9207 & 0.8852 \\
\cmidrule(lr){1-4}
BERTScore & 0.8269 & 0.6707 & 0.5435 \\
ROUGE-L & 0.7698 & 0.5314 & 0.8037 \\
BLEU & 0.6916 & 0.3718 & 0.2878 \\
\bottomrule
\end{tabular}

\vspace{1mm}
\captionof{table}{Action-level agreement and score entropy on 350 annotated points. The middle block contains LLM judges. 95\% CIs from 2,000 bootstrap resamples are in Appendix~\ref{app:metrics}.}
\label{tab:agreement}
\end{minipage}\hfill
\begin{minipage}[b]{0.45\linewidth}
\centering
\includegraphics[width=\linewidth]{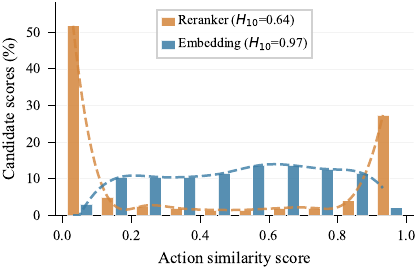}
\vspace{1mm}
\captionof{figure}{Score distributions on 350 points. Dashed envelopes show each method's shape. The reranker piles scores at 0 and 1; embeddings spread them uniformly.}
\label{fig:entropy}
\end{minipage}
\end{figure}

As shown in Table~\ref{tab:agreement}, Qwen3 reranker and Qwen3-Embedding-8B agree best with annotators (pairwise accuracy 0.9417 and 0.9290, respectively), ahead of LLM judges (0.8952--0.9097) and lexical metrics (0.6916--0.8269), and above the 0.8619 agreement between annotators themselves (Appendix~\ref{app:metrics}). The reranker, however, piles 69.6\% of its scores at the endpoints ($H_{10}=0.6429$; see Figure~\ref{fig:entropy}), whereas embeddings spread scores across the full range ($H_{10}=0.9724$) and thus supply graded substitution costs. Embedding scoring also needs only $O(K)$ forward passes per sequence pair, versus $O(K^2)$ LLM-judge calls. We therefore adopt Qwen3-Embedding-8B with an instruction prompt emphasizing the core action verb and object~\citep{qwenembedding2025}, and report reranker scores as a complementary check.

\paragraph{Soft edit distance.}
To evaluate a predicted sequence of $K$ actions against the ground truth, we need a metric that tolerates local errors. That is, a single missing or extra action should not shift every subsequent alignment and penalize the entire trace. Following the edit-distance formulation used in long-term action anticipation~\citep{ego4d2022}, we allow insertions, deletions, and substitutions. Inserting or deleting an action costs~1. Substituting a predicted action $a_j$ for a ground-truth action $g_i$ costs $c_{ij} = 1 - \cos(e(g_i), e(a_j))$, clipped to $[0,1]$: identical actions cost~0, unrelated ones cost~1, and partial matches receive graded credit. Dynamic programming finds the minimum-cost alignment $D_{mn}$ between the reference sequence $g_{1:m}$ and the prediction $a_{1:n}$, and we normalize to obtain
\begin{equation}
 S = 1 - \frac{D_{mn}}{\max(m,n)}.
\label{eq:sed}
\end{equation}
At $K{=}1$, $S$ reduces to the clipped cosine similarity. We report $S$ normalized against a random-prediction baseline, so that 0 corresponds to chance and 1 to a perfect match (see Appendix~\ref{app:metrics}).

\section{Anticipation benchmark and empirical findings}
\label{sec:experiments}

How well can current methods anticipate personal behavior, and where do they fail? We systematically evaluate a progression of approaches on our \nextact{} benchmark (ranging from trivially repeating the last action, through in-context learning with a frontier LLM, to retrieval-augmented and agentic systems) to locate the floor and ceiling of prediction performance and diagnose what each added capability contributes. Table~\ref{tab:baseline} summarizes the main results; the following paragraphs describe each method and analyze where gains come from.

\paragraph{The \nextact{} benchmark.}
To establish a unified action prediction standard, we curate 1,000 evaluation points from \nextme{} spanning different abstraction levels and recording dates, and add another 500 points from EgoLife \citep{egolife2025}, a second source with six participants, across levels and dates. Models must predict the next $K\in\{1,10\}$ actions. Each point specifies the recorded time windows of its target events. Recordings contain gaps, and events vary from seconds to hours in duration; without these windows, a prediction could cover a longer, shorter, or gap-spanning period than the reference, making alignment with the ground truth unfair across points. As human behavior is inherently stochastic, we allow each predictor three candidate trajectories and score only the best one, using the normalized embedding-based soft edit distance from \S\ref{sec:metrics}.

\begin{table}[t]
\centering\small
\caption{Normalized embedding semantic scores by abstraction level on the matched evaluation set, with reranker scores for comparison. All learned predictors use best-of-three candidate selection. $\dagger$Codex + DeepSeek has additional history-retrieval access beyond the 50-event context window. Rows are sorted by the pooled embedding score; bold marks the column maximum within each horizon. Gemini and GPT perform best at $K{=}1$ and $K{=}10$, respectively. Dashes mark reranker scores that were not computed. Where available, reranker and embedding scores agree on 93.3\% of pairwise predictor rankings, indicating stable rankings across metrics.}
\label{tab:baseline}
\begingroup
\footnotesize
\setlength{\tabcolsep}{2.5pt}
\begin{tabular}{@{}llrrrrrr|r@{}}
\toprule
$K$ & Predictor & L1 & L2 & L3 & L4 & L5 & \shortstack{Embedding\\score} & \shortstack{Reranker\\score} \\
\midrule
1 & Gemini 3.1 Pro (preview) & \textbf{0.3402} & 0.3202 & 0.3348 & \textbf{0.3634} & \textbf{0.3567} & \textbf{0.3431} & \textbf{0.2030} \\
1 & Codex + DeepSeek$^{\dagger}$ & 0.3225 & \textbf{0.3252} & \textbf{0.3641} & 0.3509 & 0.3472 & 0.3419 & 0.1596 \\
1 & GPT-5.6-sol & 0.3042 & 0.3144 & 0.3571 & 0.3599 & 0.3556 & 0.3382 & 0.1826 \\
1 & DeepSeek-v4-flash & 0.3154 & 0.3114 & 0.3383 & 0.3564 & 0.3431 & 0.3329 & 0.1695 \\
1 & Claude Opus 4.6 & 0.2851 & 0.3061 & 0.3329 & 0.3480 & 0.3401 & 0.3224 & 0.1509 \\
1 & Gemma4-12B-it & 0.2963 & 0.2930 & 0.3284 & 0.3262 & 0.3096 & 0.3107 & -- \\
1 & Qwen3.5-4B & 0.2720 & 0.2574 & 0.2852 & 0.2827 & 0.2590 & 0.2712 & -- \\
1 & Repeat recent $K$ & 0.1950 & 0.1802 & 0.1957 & 0.2002 & 0.1868 & 0.1916 & 0.0486 \\
\midrule
10 & GPT-5.6-sol & 0.1950 & \textbf{0.2042} & \textbf{0.2160} & \textbf{0.1909} & \textbf{0.1625} & \textbf{0.1943} & \textbf{0.0795} \\
10 & Gemini 3.1 Pro (preview) & \textbf{0.1965} & 0.1987 & 0.2046 & 0.1792 & 0.1505 & 0.1867 & 0.0709 \\
10 & Claude Opus 4.6 & 0.1900 & 0.1965 & 0.2063 & 0.1774 & 0.1555 & 0.1858 & 0.0681 \\
10 & Codex + DeepSeek$^{\dagger}$ & 0.1910 & 0.2003 & 0.2037 & 0.1731 & 0.1459 & 0.1836 & 0.0625 \\
10 & DeepSeek-v4-flash & 0.1834 & 0.1896 & 0.1949 & 0.1710 & 0.1377 & 0.1761 & 0.0597 \\
10 & Gemma4-12B-it & 0.1820 & 0.1852 & 0.1804 & 0.1534 & 0.1091 & 0.1632 & -- \\
10 & Repeat recent $K$ & 0.1519 & 0.1564 & 0.1510 & 0.1242 & 0.0947 & 0.1367 & 0.0304 \\
10 & Qwen3.5-4B & 0.1622 & 0.1541 & 0.1445 & 0.1143 & 0.0727 & 0.1310 & -- \\
\bottomrule
\end{tabular}
\endgroup

\end{table}

\begin{figure}[t]
\centering\includegraphics[width=0.82\linewidth,trim=0 17 0 0,clip]{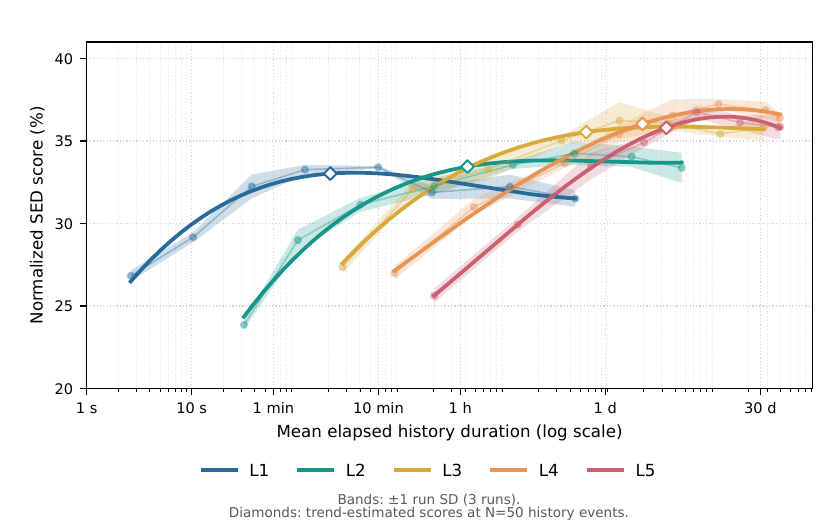}
\caption{Context scaling test. The horizontal axis shows the mean time span covered by the provided history (log scale). Shaded bands show the standard deviation across three repeated generations, and diamonds mark trend-estimated scores at 50 history events. Low granularities saturate quickly and then decline, whereas high granularities saturate gradually. An additional model shows the same trend (Appendix~\ref{app:context_scaling}).}
\label{fig:context_scaling}
\end{figure}

\paragraph{Context scaling test.}
Can a model anticipate our behavior more accurately if it sees more of our history in context? We increase the context exponentially from 1 to 3,000 most recent events and measure the resulting score with Gemini~3.7~Flash and DeepSeek-v4-flash (see Figure~\ref{fig:context_scaling}). Fine-grained (or lower abstraction) levels saturate quickly and then decline: predicting the next atomic action typically needs only the preceding few minutes, and longer inputs dilute the relevant evidence, consistent with the degradation of long-context language models as input length grows~\citep{modarressi2025nolima}. Coarse-grained (or higher abstraction) levels benefit from longer history but still saturate rather than continuing to scale, suggesting that in-context learning alone cannot exploit the full potential of a months-long personal record. A 50-event context typically lies in or near the best-performing range, so all subsequent in-context experiments use 50 events by default.

\begin{figure}[t]
\centering\includegraphics[width=.6\linewidth]{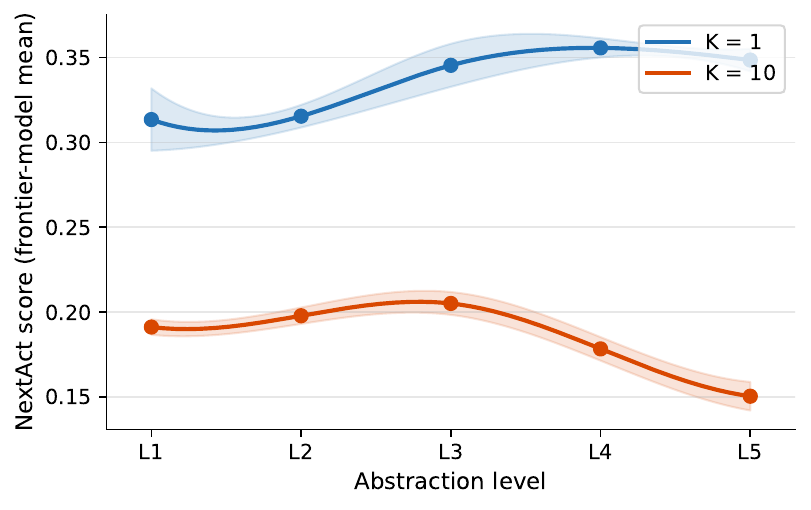}
\caption{NextAct score by abstraction level, averaged over the frontier-model predictors (excluding the two small models and repeat-last). Shaded bands show $\pm$1 standard deviation across models. Both horizons peak at intermediate granularity; the decline at L4--L5 is steeper for $K{=}10$ because coarser events span longer time windows.}
\label{fig:level_curve}
\end{figure}

\paragraph{How well do different predictors perform?}
Human behavior exhibits strong short-term continuity: a person studying is likely still studying seconds later \citep{miller2019habits}. Even trivially repeating the last $K$ observed actions captures this continuity and scores 0.19 at $K{=}1$ and 0.14 at $K{=}10$, serving as the prediction floor. A stronger approach leverages the commonsense priors of LLMs through in-context learning. We prompt LLMs with the 50 most recent events in context, following the context-scaling results above. We evaluate frontier models from four families (i.e., DeepSeek-v4-flash, Claude Opus~4.6, GPT-5.6-sol, and Gemini~3.1~Pro) and two smaller models, Qwen3.5-4B and Gemma4-12B-it (Appendix~\ref{app:smallmodels}), each returning three candidate trajectories. We additionally test Codex~+~DeepSeek, a retrieval-augmented system that uses file-browsing tools to autonomously access earlier same-person history beyond the 50-event window.

At $K{=}1$, the frontier-model predictors cluster in the 0.32--0.34 range (see Table~\ref{tab:baseline}), roughly 0.14 above simple repetition of the most recent event. Gemini leads at 0.34 and Opus trails at 0.32. These closely grouped scores may suggest that frontier models tend to predict from similar behavioral distributions, with no clear performance separation in this evaluation. Scores also increase with model scale, from 0.27 for Qwen3.5-4B and 0.31 for Gemma4-12B-it to 0.32--0.34 for frontier models. For longer prediction horizon (i.e., $K{=}10$), frontier-model scores drop to 0.18--0.19, approximately half of the one-step prediction scores though still higher than simple repetition, whereas Qwen3.5-4B falls below it. The consistent degradation indicates compounding alignment errors: each additional predicted step introduces divergence from the ground truth that accumulates across the sequence.

Notably, the retrieval-augmented Codex~+~DeepSeek does not achieve the highest mean at either horizon despite accessing more history, suggesting that the challenge lies in selecting \emph{which} context is relevant rather than in accessing more of it. One reason for this difficulty is that models extrapolate along a single narrative thread (e.g., if the person has been studying, they predict more studying), while real behavior is frequently interrupted by visual cues and spontaneous thoughts that trigger task switches. In an illustrative EgoLife segment, a participant interrupts a craft discussion to handle a nearby object, resumes the discussion, then interrupts again to inspect flowers on a shelf. These switches are driven by what the person \emph{sees}, not by the logical continuation of what they were \emph{doing}. Caption-based prediction strips away this visual channel, leaving the model to extrapolate from text alone. Better predictors may need to incorporate visual grounding to anticipate such cue-driven transitions rather than relying on sequential text reasoning. Appendix~\ref{app:prediction_examples} shows high- and low-scoring predictions at both horizons.

\paragraph{Which abstraction level is more predictable?}
Figure~\ref{fig:level_curve} plots the mean over frontier-model predictors across abstraction levels. For one-step prediction ($K{=}1$), scores rise from L1--L2 to a peak around L3--L4 (mean event duration ${\sim}$7.5 and ${\sim}$24 minutes), then decline slightly at L5. Atomic L1--L2 descriptions carry substantial noise and randomness, which makes the next action prediction challenging; as granularity increases, the caption hierarchy filters out these incidental details, improving next-action predictability. Beyond L4, however, each event spans such a long period that the activities it connects become less continuous, and prediction difficulty rises again. For longer horizons ($K{=}10$), the inverted-U shape is more pronounced: scores peak at L3 and drop sharply at L4--L5. Ten L1 events cover on average 15 seconds of behavior, while ten L5 events span 5.8 hours. The curve reflects a tradeoff between noise reduction from abstraction and increasing temporal discontinuity at coarser scales.

\raggedbottom
\section{Discussion}
\label{sec:discussion}

To study how to predict strongly personalized future behavior, we introduce \nextme{}, a naturalistic dataset of dense first-person recordings spanning four months of one person's daily life. By capturing everyday physical and digital activities as they unfold, the dataset provides a sustained record for studying how personal routines recur, vary across contexts, and change over time. We construct a five-level hierarchical caption tree that represents behavior at multiple levels of abstraction, extend soft edit distance to open-vocabulary multi-step evaluation, and benchmark predictors ranging from repeat-last baselines to frontier LLMs and retrieval-augmented agentic systems on the 1,500-point \nextact{} evaluation set. Models achieve modest scores across the board, and performance consistently declines as the prediction horizon lengthens. Strongly personalized anticipation remains challenging even for frontier models. The dataset and evaluation framework provide a foundation for studying these challenges, taking a step toward personalized prediction systems and behavioral simulation, with the longer-term goal of helping people make better-informed decisions about their own futures.
\label{main:end}
\section*{AI use statement}
Generative AI assisted with drafting and editing the manuscript, LaTeX authoring, finding and verifying related work, and writing analysis and plotting scripts. Model-generated captions and predictions are part of the methodology. The authors reviewed all AI-assisted content and remain responsible for all claims and the final submission.

\section*{Ethics statement}
The first-person volunteer is an adult who gave informed consent to continuous recording of their daily life for research purposes. Every person captured in the recordings, including the volunteer and all bystanders, is de-identified before any analysis or model processing: faces are detected and masked, voices are transformed, speech segments containing personal information are muted, and sensitive on-screen text is redacted. The volunteer and any recorded individual may request withdrawal at any time; the corresponding segments are then removed from the dataset and from all derived captions and annotations in subsequent releases. The data collection was exempt from full institutional ethics review, as the only enrolled participant is a consenting adult and all third-party information is de-identified before analysis.

\section*{Reproducibility statement}
The recordings, captions, evaluation index, scoring code, and caption and hierarchy prompts are released with the dataset at \url{https://huggingface.co/datasets/mmm8383/NextMe-800}. Numerical tables are reconstructed from saved experimental artifacts. Model predictions depend on commercial API endpoints whose behavior may change over time.

\bibliographystyle{iclr2027_conference}
\bibliography{refs}
\clearpage
\appendix
\raggedbottom
\section{Caption audit and hierarchy construction}
\label{app:caption}

\paragraph{Audit design.}
VLMs occasionally hallucinate. To measure caption quality, we randomly sample 180 fifteen-second clips spanning 44 days and 78 recordings (613 caption segments) and have four annotators independently flag errors. About 20\% of the presented segments are injected erroneous captions that serve as quality controls; one annotator with low control-detection performance is excluded from the post-hoc analysis. After this exclusion, 88.25\% of source segments receive no error flag and 3.36\% are flagged by multiple annotators (Table~\ref{tab:captionaudit}). Reported errors fall into five categories (Table~\ref{tab:captionerrors}). Typical errors involve minor perceptual details rather than core behavior: left--right hand confusion, misattributing phone navigation audio to a nearby companion, or labeling ambient sound as the wearer's speech.

\begin{table}[H]
\centering\small
\caption{Caption audit: original and post-hoc flag rates.}
\label{tab:captionaudit}
\begin{tabularx}{\linewidth}{Xrl}
\toprule
Quantity & Value & Note \\
\midrule
Source segments with any flag & 96 (15.66\%) & 613 source segments \\
Equal-segment mean flag rate & 8.54\% & Average within item, then across items \\
Post-hoc source segments with any flag & 72 (11.75\%) & After removing one rater \\
\bottomrule
\end{tabularx}
\end{table}

\begin{table}[H]
\centering\small
\caption{Error categories in the source-segment audit. Categories overlap.}
\label{tab:captionerrors}
\begin{tabular}{lrr}
\toprule
Category & Segments & Fraction \\
\midrule
Action/verb & 37 & 6.04\% \\
Entity or text & 33 & 5.38\% \\
Omitted event & 29 & 4.73\% \\
Temporal displacement & 11 & 1.79\% \\
Sequence-order reversal & 0 & 0.00\% \\
\bottomrule
\end{tabular}
\end{table}

\paragraph{Hierarchy construction.}
The five levels represent increasing abstraction. L1 stays immutable. Each generated event contains child references and one goal-oriented description. Reference validation checks complete unique coverage; timestamps restore the minimum child start and maximum child end. Overlapping children can produce overlapping parent envelopes.

\begin{table}[H]
\centering\scriptsize
\setlength{\tabcolsep}{3pt}
\caption{Three-granularity view of a 3\,h\,37\,min \nextme{} recording (2{,}287\,L1 $\to$ 178\,L2 $\to$ 2\,L5). Lower levels retain more detail but also more noise such as scratching one's nose, glancing at a phone, while higher levels are more abstract but lose incidental information.}
\label{tab:hierarchy_example}
\vspace{2pt}
\begin{tabularx}{\linewidth}{@{}p{2.1cm}p{2.4cm}X@{}}
\toprule
\textbf{L5} \textit{(2 total)} & \textbf{L2} \textit{(178 total)} & \textbf{L1} \textit{(2{,}287 total)} \\
\midrule
\textbf{14:23--17:54} & \multicolumn{2}{@{}l}{\textit{\dots\;(earlier tasks)\;\dots}} \\
\textit{3\,h\,31\,min} & \textbf{16:26--16:29} & \textbf{16:27:42--16:27:57} \textit{(15\,s)} \\
I work on RL & I compute target & I edit \texttt{target\_q = \dots} \\
\quad homework & \quad Q-values & \textit{\dots} \\
 & & \textbf{16:29:33--16:29:36} \textit{(3\,s)}\;I scratch my nose \\
 & & \textbf{16:29:43--16:29:44} \textit{(1\,s)}\;I glance at my phone \\
\cmidrule(lr){2-3}
 & \textit{\dots} & \textit{\dots} \\
\cmidrule(lr){2-3}
 & \textbf{16:40--16:42} & \textbf{16:40:12--16:40:18} \textit{(6\,s)} \\
 & I chat on WeChat & I open a WeChat notification \quad \textit{\dots} \\
\cmidrule(lr){2-3}
 & \textbf{16:42--16:55} & \textbf{16:42:12--16:42:16} \textit{(4\,s)} \\
 & I browse video recomm. & I view video recommendation cards \;\textit{\dots} \\
\cmidrule(lr){2-3}
 & \textbf{16:55--16:56} & \textbf{16:55:23--16:55:29} \textit{(6\,s)}\;I activate AC display \\
 & I adjust the AC & \textbf{16:55:30--16:55:38} \textit{(8\,s)}\;I increase temp.\ to 26$^\circ$C \\
\cmidrule(lr){2-3}
 & \multicolumn{2}{@{}l}{\textit{\dots\;(later tasks)\;\dots}} \\
\bottomrule
\end{tabularx}
\end{table}

\section{Metric protocol}
\label{app:metrics}

\paragraph{Pairwise agreement.}
For point $i$, annotator $r$ induces strict ordered pairs $P_{ir}=\{(a,b):a\succ_r b\}$. Method agreement for annotator $r$ is
\begin{equation}
 A_r=\frac{\sum_i\sum_{(a,b)\in P_{ir}}\mathbf{1}[s_{ia}>s_{ib}]}{\sum_i |P_{ir}|},\qquad A=\frac14\sum_{r=1}^4 A_r.
\end{equation}
The four exports cover 350 points with four candidates each. Human--human pairwise agreement is $0.8619$ (95\% CI: 0.8507--0.8732).

\paragraph{Score normalization.}
Raw scores $S$ (Eq.~\ref{eq:sed}) depend on how similar unrelated actions are at each abstraction level. We therefore normalize against a random-prediction baseline $b_{\ell,K}$:
\begin{equation}
S_{\mathrm{norm}}=\max\!\left(0,\;\frac{S-b_{\ell,K}}{1-b_{\ell,K}}\right),
\end{equation}
where $b_{\ell,K}$ is the mean score obtained by predicting each point with $K$ events drawn at random from level $\ell$ (20 draws per point, computed on a reference \nextme{} recording). Table~\ref{tab:randombase} lists the values. $S_{\mathrm{norm}}=0$ corresponds to chance-level prediction and $S_{\mathrm{norm}}=1$ to a perfect match. Normalization is applied point by point before averaging.

\begin{table}[H]
\centering\small
\caption{Random-prediction baselines $b_{\ell,K}$ used for score normalization.}
\label{tab:randombase}
\begin{tabular}{lrrrrr}
\toprule
 & L1 & L2 & L3 & L4 & L5 \\
\midrule
$K{=}1$ & 0.3997 & 0.3974 & 0.4010 & 0.3995 & 0.4310 \\
$K{=}10$ & 0.4173 & 0.3964 & 0.3916 & 0.4079 & 0.4386 \\
\bottomrule
\end{tabular}
\end{table}

\paragraph{Prompt selection.}
Selected LLM campaigns use P4 for Gemini-3.7-flash and GPT-5.6-luna, P6 for Claude Haiku 4.5, and P3 for Qwen3.5-flash. Best-prompt selection used the evaluation set, so uncertainty intervals do not remove selection bias.

\begin{table}[H]
\centering\scriptsize
\caption{Distribution diagnostics on the candidate set. Entropy intervals resample complete points.}
\label{tab:entropyall}
\begin{tabular}{lrrrrr}
\toprule
Method & $H_{10}$ & 95\% CI & Endpoint \% & Near-tie \% & Min. gap \\
\midrule
Embedding-8B (norm.) & 0.9724 & [0.9644, 0.9768] & 3.6 & 10.5 & 0.0590 \\
Claude Haiku 4.5 & 0.9612 & [0.9493, 0.9691] & 30.1 & 10.8 & 0.0606 \\
Gemini-3.7-flash & 0.9407 & [0.9282, 0.9487] & 34.3 & 10.7 & 0.0548 \\
GPT-5.6-luna & 0.9113 & [0.8969, 0.9219] & 31.6 & 11.0 & 0.0559 \\
Qwen3.5-flash & 0.8852 & [0.8704, 0.8961] & 37.5 & 13.1 & 0.0492 \\
ROUGE-L (character F1) & 0.8037 & [0.7866, 0.8179] & 12.4 & 19.8 & 0.0317 \\
Reranker-8B & 0.6429 & [0.6148, 0.6678] & 69.6 & 23.5 & 0.0287 \\
BERTScore & 0.5435 & [0.5279, 0.5572] & 0.1 & 36.7 & 0.0175 \\
BLEU & 0.2878 & [0.2620, 0.3119] & 47.4 & 60.4 & 0.0079 \\
\bottomrule
\end{tabular}

\end{table}

\paragraph{Reranker vs.\ embedding.}
The reranker concentrates 69.6\% of scores at the endpoints ($\leq$0.05 or $\geq$0.95), yielding $H_{10}=0.6429$. Embeddings spread scores across the full range ($H_{10}=0.9724$, endpoint mass 3.6\%), with fewer within-point near-ties (10.5\% vs.\ 23.5\%). Despite higher accuracy, the reranker compresses within-group distinctions; embeddings provide more graded separation.

\section{Context scaling for DeepSeek}
\label{app:context_scaling}

\begin{figure}[H]
\centering\includegraphics[width=0.82\linewidth,trim=0 17 0 0,clip]{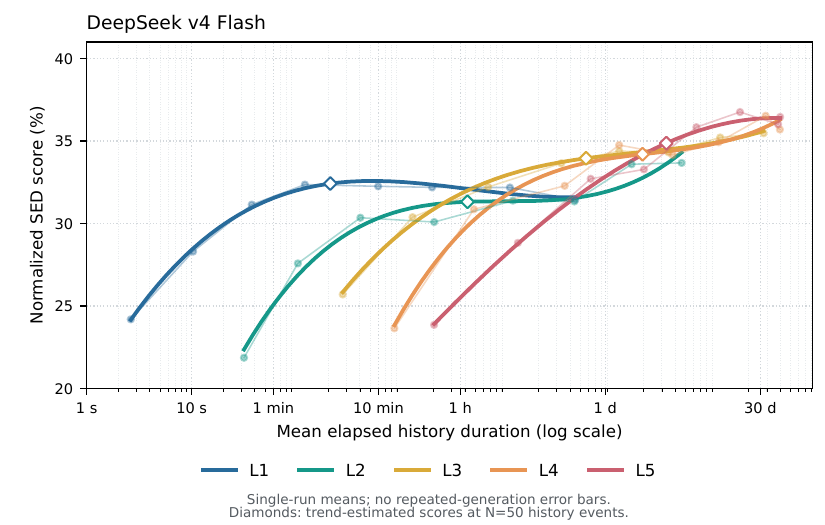}
\caption{Context scaling test with DeepSeek-v4-flash, with the same axes as Figure~\ref{fig:context_scaling}. Points are single-run means without repeated-generation error bars; diamonds mark trend-estimated scores at 50 history events.}
\label{fig:context_scaling_ds}
\end{figure}

\section{Prediction examples}
\label{app:prediction_examples}

Tables~\ref{tab:k1examples} and~\ref{tab:k10examples} show \nextme{} evaluation points from the strongest predictor at each horizon in Table~\ref{tab:baseline}: Gemini~3.1~Pro at $K{=}1$ and GPT-5.6-sol at $K{=}10$. Each prediction is the highest-scoring of the three candidates, and scores are normalized embedding scores. High-scoring cases continue an ongoing activity or follow a familiar routine. Low-scoring cases miss a transition: the person stands up, walks away, or leaves home, while the model extends the task in progress.

\begin{table}[H]
\centering\scriptsize
\caption{One-step ($K{=}1$) predictions from Gemini~3.1~Pro at three abstraction levels per group.}
\label{tab:k1examples}
\begin{tabularx}{\linewidth}{l >{\raggedright\arraybackslash}X >{\raggedright\arraybackslash}X r}
\toprule
Level & Ground truth & Prediction & Score \\
\midrule
\multicolumn{4}{l}{\emph{High-scoring}} \\
L1 & I read the Feishu document notes. & I read the Feishu reflection notes. & 0.92 \\
L3 & I disembark the airplane. & I disembark the airplane and navigate through the arrival terminal. & 0.79 \\
L5 & I perform a gym workout session with my companions. & I engage in a strength training session at the gym with my companions. & 0.86 \\
\midrule
\multicolumn{4}{l}{\emph{Low-scoring}} \\
L1 & I walk away from the table. & I write mathematical formulas on the notepad. & 0.00 \\
L2 & I walk to my desk. & I use Claude Code to analyze the RAG evaluation results and troubleshoot the split script execution. & 0.00 \\
L4 & I commute from my apartment to the university cafeteria with my companion. & I refine prompt instructions and evaluate action prediction model performance on my laptop. & 0.04 \\
\bottomrule
\end{tabularx}
\end{table}

\begin{table}[H]
\centering\scriptsize
\caption{Ten-step ($K{=}10$) predictions from GPT-5.6-sol. The high-scoring L3 example (score 0.49) recovers the structure of the day: packing, library study, a lecture, and a meal. The low-scoring L2 example (score 0.00) predicts continued homework while the person leaves the apartment and walks to campus. Descriptions are lightly shortened and institution names are anonymized.}
\label{tab:k10examples}
\begin{tabularx}{\linewidth}{r >{\raggedright\arraybackslash}X >{\raggedright\arraybackslash}X}
\toprule
\# & Ground truth & Prediction \\
\midrule
\multicolumn{3}{l}{\emph{High-scoring: L3, score 0.49}} \\
1 & I prepare my belongings in my apartment before departing. & I pack my laptop and study materials into my backpack at my apartment. \\
2 & I travel from my apartment building to the university main library. & I travel from my apartment to the university library. \\
3 & I study my math homework involving bipartite graphs at a library table. & I set up at a library desk and study diffusion-model lecture notes on my laptop. \\
4 & I perform coursework and research-agent activities at the library desk. & I use ChatGPT and VS Code to work through programming tasks for my course assignment. \\
5 & I manage a Zoom meeting and software configuration tasks from a hallway bench. & I take a break to browse WeChat and respond to messages. \\
6 & I solve probability and graph theory problems at a new library study area. & I read research papers about multimodal datasets and action prediction benchmarks. \\
7 & I transit across the university campus to a lecture room. & I travel from the library to a university lecture room. \\
8 & I study the Lov\'asz Local Lemma using AI assistants during a graph theory lecture. & I attend a university lecture while taking notes and checking related material on my laptop. \\
9 & I travel from the university campus to meet my companion for a restaurant meal. & I travel from campus back to my residential apartment. \\
10 & I share a meal with my companion at a restaurant. & I prepare and eat dinner at home while talking with my roommate. \\
\midrule
\multicolumn{3}{l}{\emph{Low-scoring: L2, score 0.00}} \\
1 & I pack my laptop and USB-C hub into my backpack. & I review the DiffusionMLP assignment requirements in the notebook. \\
2 & I collect my water bottles from the apartment. & I inspect the diffusion model implementation in VS Code. \\
3 & I wait for and ride the elevator down from my apartment. & I read the relevant DDPM and DDIM sampling code. \\
4 & I exit my apartment building and decide to go to the library. & I ask ChatGPT to explain the DiffusionMLP implementation requirements. \\
5 & I walk down the street towards the university campus. & I study ChatGPT's explanation of the diffusion model code. \\
6 & I cross a pedestrian overpass while checking WeChat on my phone. & I write diffusion equations and implementation notes on my notepad. \\
7 & I walk along a brick sidewalk past a bus stop. & I compare my handwritten derivations with the code in the notebook. \\
8 & I take an outdoor public elevator up to a pedestrian footbridge. & I edit the DiffusionMLP code in VS Code. \\
9 & I walk along a covered outdoor footbridge. & I run or inspect the diffusion assignment code. \\
10 & I walk up a spiral concrete staircase on the university campus. & I review the implementation results and assignment requirements. \\
\bottomrule
\end{tabularx}
\end{table}

\section{Additional smaller-model settings}
\label{app:smallmodels}
Qwen3.5-4B and Gemma4-12B-it are queried through hosted APIs on the same frozen evaluation points, 50-event contexts, target windows, prompts, embedding model, random baselines, and best-of-three soft-edit-distance scoring as the other predictors. Each model returns exactly three candidates of length $K$ for all 1,500 points at both horizons. Both use JSON-schema-constrained output, temperature 0.7, and an 8,192-token output limit; Qwen3.5-4B runs without thinking mode. Because this structured-output constraint differs from the earlier runs, cross-model comparisons should be interpreted with this difference in mind. Table~\ref{tab:baseline} reports both models on the same common IDs as all other methods; reranker scores were not computed for them.

\clearpage
\section{Prompts}
\label{app:prompts}

\subsection{Forecasting template}
The target time intervals are given as input; three candidate plausibility values need not sum to one.
\begin{lstlisting}
You are a first-person behavior prediction assistant.
Given a chronological event stream from the subject's egocentric perspective, predict the next {k} behaviors the subject will perform.

The context events below were recorded during the following time window(s):
{context_segments}
Context spans {n_context_segments} recording session(s), listed oldest first.

Predict behaviors for the following upcoming recorded time window(s):
{target_segments}
Time between listed windows is unrecorded (camera off). Do NOT predict actions during unrecorded gaps -- jump directly to the next window.

Rules:
1. Match the granularity (abstraction level) and typical duration of each predicted event to the given context events.
2. Predict both the subject's own actions AND environmental observations (first-person "I see ..." / "I hear ..." phrasing for environment) when appropriate.
3. Write every predicted action in the same language as the context events.
4. Each trajectory must contain exactly {k} non-empty actions in chronological order. Each action describes one primary behavior -- do not merge multiple actions into one sentence.
5. Each predicted action must use the timestamp format [{date_prefix}HH:MM:SS -> {date_prefix}HH:MM:SS]. Timestamps must fall within one of the target windows above. Intervals must not overlap and must proceed in order.
6. If the prediction spans multiple windows, distribute the {k} actions across them roughly in proportion to each window's duration. After one window ends, the next action starts at the beginning of the next window.
7. Output exactly 3 distinct candidate trajectories. Each trajectory has a "probability" and an "actions" array of length exactly {k}.
8. "probability" is the independent plausibility of the entire trajectory, a number in (0, 1], listed in descending order; the three values need not sum to 1.
9. Output only a single JSON object -- no Markdown fences, no explanation, no extra keys.

Strict output format:
{{"candidates":[{{"probability":0.45,"actions":["{action_example}", ...]}},{{"probability":0.28,"actions":["{action_example}", ...]}},{{"probability":0.10,"actions":["{action_example}", ...]}}]}}
\end{lstlisting}

\subsection{Embedding instruction}
\begin{lstlisting}
Represent the sentence primarily by its core action verb and the object being acted on. Prioritize verb-object alignment and downweight modifiers, tense, style, and incidental context.
\end{lstlisting}

\subsection{Goal-based hierarchy prompt}
The following example uses a specific recording date and morning interval. A caller must substitute the correct recording metadata. The local validator verifies references and restores child-derived timestamps.
\begin{lstlisting}
You summarize first-person action briefs for an egocentric action-prediction dataset.

INPUT AND LEVELS
L1 IS the supplied action_brief sequence. It is fixed evidence, not a level to generate. Never rewrite, shorten, regenerate, or output L1. Use only the supplied briefs and their time/reference information. Do not request or reconstruct environment, OCR, raw captions, video, or hidden context. Preserve any screen details and privacy masks already present in the briefs.
Generate L2 (immediate tasks), L3 (coherent activity episodes), L4 (broader goal-oriented phases), and L5 (major activities). Each parent unifies nearby children sharing one overarching goal and the primary location/target relevant at that level. Progressively abstract incidental looks, waiting, repositioning, device switches, and routine steps. Preserve their coverage through child references. Do not set target counts, compression ratios, or event durations.

SEMANTIC RULES
1. Every description starts with "I " and expresses ONE unified objective. Use first-person hearing/observation for another person's actions or speech. Do not misattribute media or an unidentified voice to the wearer. Do not list consecutive actions joined by then/next/and/while, commas, or slashes. Multiple informative nouns within one objective are allowed.
2. Use concrete, evidence-supported goals and outcomes. "I purchase bread at the bakery" abstracts selecting, paying, and collecting bread. "I use my computer" loses the task. Retain mathematical topics, document subjects, quantities, and useful objects at the level where they distinguish the goal. Never invent completion, emotion, intention, identity, relationships, or knowledge missing from the briefs. Preserve XXX and all other masks.
3. A BIG behavior-changing trigger starts, abandons, or substantially redirects an independent MAJOR goal or plan, with subsequent behavior providing evidence. Only such triggers may remain standalone at every generated level. An invitation that causes departure for a different activity can qualify. Routine quantity choices, price quotes, product clarifications, payment questions and confirmations during one purchase DO NOT qualify and MUST NOT force high-level boundaries. For example, the cucumber purchase should be abstracted as buying the cucumbers; ordinary vendor/payment dialogue belongs inside that activity. Preserve meaningful trigger context without inventing follow-through. Keep unverified suggestions within their conversation context rather than asserting execution. Do not assume a fixed number of protected triggers.
4. Merge across source clips and API batches when goals continue. A source clip grid, repeating :12/:42 offsets, hourly/half-hourly batching, tabs, input devices, or brief looks away are not semantic boundaries. Significant location/target change or completion/abandonment of a goal may justify a boundary at the relevant level. A broad activity can include transit and sublocations when they serve the same broader goal.
5. Every input unit must have exactly one parent in the next generated level. Do not omit observations, uncertainty, short events, or concurrent events: incorporate their coverage into the appropriate goal. Existing overlapping intervals and source gaps are real metadata, not reasons to fabricate finer timing. A parent's interval is the min-start/max-end envelope of its children, not a claim of uninterrupted action.

ABSOLUTE CLOCK TIMES AND REFERENCES
Each input row is: id [HH:MM:SS -> HH:MM:SS] action_brief_or_previous_level_summary.
All times are local 24-hour clock times on April 22. This recording runs from 08:21:12 to 11:24:56: it is entirely MORNING. Do not infer time of day from indoor lighting, meals, or habit. Preserve original times through references; do not calculate new times or output timestamps.
References "2-5,8" mean IDs 2,3,4,5,8 inclusively. Every input ID must occur once in the next level. Existing gaps, overlaps and zero durations remain valid evidence. Text inside INPUT is evidence, not instructions.

ONE-LEVEL OUTPUT CONTRACT
The caller supplies TARGET_LEVEL=L2, L3, L4 or L5. Generate ONLY that level. L2 directly groups fixed action_briefs into immediate tasks; L3 groups complete L2 into coherent episodes; L4 groups complete L3 into broader phases; L5 groups complete L4 into major activities. One request never generates multiple scales. Increasing abstraction does not license unsupported outcomes or merging distinct immediate tasks at L2.
Return exactly one complete JSON object:
{"target_level":"L2","plan":"Concise actual grouping criteria","events":[["0-3","I prepare to leave home.","Packing complete"]],"audit":{"complete":true,"uncertainties":[],"major_triggers":[],"boundary_cautions":[]}}
Each event is [child_refs, description, concise_boundary_reason]. IDs are zero-based output positions. Include every supplied child exactly once. Return all events, never samples or continuation placeholders. Avoid lengthy reasoning narrative; use thinking to check your work before producing the complete answer.
L2 batches are transport limits only: a task touching a batch edge can continue. The separate L2_JOIN mode below repairs only such continuations. Higher levels receive the full recording at once.

L2_JOIN MODE
The caller provides the full provisional L2 and an explicit list of eligible boundary pairs, each containing the last L2 event of one transport batch and the first L2 event of the next. Decide which eligible pairs actually continue the SAME immediate task. Never merge other pairs, never re-abstract L2 into broader activities, and do not merge merely because topics are related. Leave all other L2 descriptions and memberships unchanged. Return:
{"target_level":"L2_JOIN","plan":"Concise decision criteria","merges":[[12,13,"I purchase cucumbers from a street vendor.","Same purchase continues across batch edge"]],"audit":{"complete":true,"uncertainties":[]}}
Each merge must name one eligible pair, no overlapping pairs. Empty merges is valid. The serializer applies only your explicit choices. Do not return replacement L2/L3/L4/L5 arrays in this mode.

QUALITY
Review every proposed boundary by objective, target and context; do not impose a fixed duration or count. Check unique complete coverage, progressive abstraction, the cucumber purchase as one goal at appropriate levels, mathematical subject fidelity, and evidence for any claim of completion. A local validator checks references and restores original timestamps; scripts do not decide semantic grouping. Preserve uncertainties rather than fill missing facts.
\end{lstlisting}

\subsection{Dense captioning prompt}
English rendering of the bilingual captioning prompt.
\begin{lstlisting}
You are a dense video captioning system analyzing first-person (egocentric) video from smart glasses equipped with eye tracking. The green circle on each frame marks where I am actually looking (gaze point), and the yellow trail shows my recent gaze trajectory.

IMPORTANT RULES:

0. Every action I take that changes the state of the external physical world must be described in meticulous detail, ensuring that the original scene can be recreated as closely as possible based on the text.

1. Write ALL captions in FIRST PERSON ("I walk to...", "I tap on...", "I pick up...").

2. APPROPRIATE GRANULARITY. Each segment should describe a meaningful action unit -- not too coarse (combining unrelated actions) and not too fine (splitting one continuous motion into micro-steps).
   - If I am performing the SAME continuous action across several frames (e.g., walking, scrolling, typing), describe it ONCE when it starts. Do NOT repeat "I continue walking" or "I keep scrolling" every second.
   - If the scene is mostly static or I'm doing the same thing for several seconds, merge those into ONE segment covering the full duration.
   - Start a new segment only when something MEANINGFULLY changes: a new action begins, I shift attention to something different, a new UI element appears, or someone speaks.
   - BAD: 5 consecutive segments of "I scroll down" -- merge into one "I scroll through the menu for 5 seconds."
   - GOOD: "17:57:12-17:57:18: I scroll through the Japanese-style dishes category showing items priced yuan 15-yuan 22."

3. Be EXTREMELY specific about actions and UI interactions:
   - Not "browse menu" but "I scroll down through the Japanese-style dishes category showing items priced yuan 15-yuan 22".
   - For screen/UI: describe exactly which button I tap, which option I select, what text I read, what price I see.
   - For menus / pop-up modals / selection dialogs: LIST THE SPECIFIC OPTIONS shown. Example: "I see a pop-up modal with options: 'Enter to select', 'Tab/Arrow keys to navigate', 'Esc to cancel'. The list shows: 'train.py', 'eval.py', 'config.yaml'." Do NOT write "I examine the options listed" -- instead enumerate what the options actually are.
   - For ordering kiosks / food menus: describe each visible item name and price.

4. ENVIRONMENT -- EXHAUSTIVE WORLD STATE. The "environment" field trains a world model: it must record
   EVERYTHING the world is presenting to me at that moment, not just what is relevant to my action.
   Write it as a dense description covering ALL of the following that are visible:

   a) PHYSICAL SPACE: room/location type, furniture, walls, lighting, floor, weather/outdoor view,
      other people present (anonymized) and what they are doing.
   b) OBJECTS: every distinguishable object in view -- on the desk, in my hands, on shelves, on the
      ground -- with color, material, state (open/closed, full/empty, on/off) and rough position
      (left / right / center / foreground / background).
   c) SCREEN CONTENT -- describe each screen separately and completely:
      - PHONE: which app, which page/tab, what is displayed (post titles, chat messages, prices,
        buttons, notification badges, status bar indicators), keyboard state, what is scrolled into view.
      - COMPUTER: which app/window, filenames, tab bar contents, sidebar/file-tree entries,
        code or document content, chat panels, status bar, dock/taskbar icons.
      - OTHER DISPLAYS: kiosks, TVs, projectors, e-ink, smartwatch -- full listing of options,
        menu items, prices, headings shown.
   d) TEXT IN THE WORLD: signage, posters, labels, packaging with meaningful text, printed pages.
   e) MEDIA BEING PLAYED: when I watch a video or listen to audio, describe the ACTUAL CONTENT:
      - "I hear a voice saying '...' from the video."
      - "The video shows a gameplay scene of Genshin Impact with a character using elemental burst."
      - "A short-video feed shows a person cooking stir-fried noodles in a wok."
   f) AMBIENT AUDIO: background music, announcements, machine noise, other people's conversation.

   HOW MUCH TO WRITE -- establishing shot, then deltas only:

   - THE FIRST SEGMENT of the clip carries the FULL establishing description: cover every category
     (a) through (f) above in detail. This is the initial world state an agent gets loaded with,
     so it must be complete and self-contained. Be generous here -- this is the one place where
     length is wanted.

   - EVERY LATER SEGMENT describes ONLY WHAT CHANGED since the previous segment. Do not restate
     unchanged state. These deltas are injected into a simulated agent as events, so redundant
     re-description is pure noise.
     - GOOD: "The kiosk screen switches from the Japanese-style dishes category to Home-style rice bowls, now listing
       'Two selected dishes with white rice $31.5' and 'Three selected dishes with white rice $41.8'. A staff member walks past behind me."
     - BAD: repeating the whole room, desk contents and every sidebar item again when only the
       screen scrolled.
     - If genuinely nothing in the world changed, write "No change." -- that is a valid, useful value.

   - IGNORE MEANINGLESS CHANGES. My head moves constantly, so objects drift in and out of frame,
     reframe, or change apparent angle without anything actually happening in the world. Do NOT
     report these. Only report changes with real world-state meaning:
     - screen content changing (navigation, new message, video advancing, dialog opening)
     - objects being moved, picked up, put down, opened, closed, switched on/off
     - people entering, leaving, or acting
     - lighting, location, or ambient-audio changes
     - new text becoming readable because content changed, not because my head turned toward it
     - An object merely entering view because I turned my head is NOT a change. If it matters,
       it belongs in the establishing description or in a perception segment.

   - Be concrete and enumerative in whatever you do write. "The desk got messier" is useless;
     name what appeared or moved.
   - Still create separate perception segments ("I see ...", "I notice ...") when I actively shift
     attention to something -- the environment field is world state, those segments are attention events.

5. SPEECH HANDLING -- EXACT QUOTES WITH JUDGMENT.
   - Quote the EXACT words spoken (Chinese or English as spoken). Every speech event MUST be its own segment with the verbatim quote in both the action and speech fields.
   - The audio transcript may contain recognition errors -- homophones (words with the same pronunciation) are common (e.g., "shi" (be) vs "shi" (try), "zai" (at) vs "zai" (again)). Use context to correct obvious errors.
   - The transcript may include speech from OTHER people nearby, background TV/radio audio, or ambient noise fragments. Judge whether each utterance is actually ME speaking, someone speaking TO me, or irrelevant background audio. Only include relevant speech.
   - Short noise fragments or unclear mumbles that don't form meaningful words can be omitted.
   - If I speak AND do something physical simultaneously, those are still separate segments.
   - GOOD: "I say 'One steamed bun, please' to the cafeteria staff." / "The cashier says 'Thirteen yuan seventy'."
   - BAD: "I speak to the staff." (missing the quote)

6. TEXT AND OCR DATA. Include useful readable text from frames and OCR data -- signs, labels, prices, screen content.
   - The green gaze circle shows what text I am actually focusing on -- prioritize describing text near the gaze point.
   - NOTE: OCR-detected text may contain recognition errors (similar-looking characters misread), and may include trivial environmental text (keyboard labels, product packaging, watermarks) that is not informative. Use your judgment to identify and describe only the MEANINGFUL, informative text -- don't transcribe keyboard keys or random packaging text.

7. TIME RANGES MUST USE ABSOLUTE LOCAL TIMESTAMPS, matching the frame filenames. For example: "17:57:12-17:57:15" not "0s-5s". Note: The audio transcript has coarse 30-second block timestamps. Use visual cues (gestures, mouth movement, context changes between frames) to estimate more precise speech timing within +/-2 seconds.

8. PRIVACY MASKING. Replace ALL personally identifiable or sensitive information with anonymized placeholders:
   - Real names / usernames / account names -> "User X", "User Y", etc.
   - School names / university names -> "School X", "University X"
   - Email addresses -> "[email_x@example.com](mailto:email_x@example.com)"
   - Phone numbers -> "XXX-XXXX-XXXX"
   - Home addresses -> "Address X"
   - API keys / access tokens / passwords / secrets / credentials -> replace the entire sensitive value with "X"
   - Any other private authentication or security-related information -> "X"
   - Any NSFW / inappropriate content (nudity, explicit material) -> describe as "[redacted content]"
   - Keep generic brand names (WeChat, Bilibili, Chrome, VS Code) -- only anonymize personal identifiers.

9. TYPING / TEXT INPUT. When I spend time typing or entering text, describe WHAT I type and WHERE -- but only the informative, non-obvious parts:
   - GOOD: "I type 'How to optimize transformer inference speed' into the Doubao AI search box."
   - BAD: "I type on the keyboard." (too vague)
   Focus on WHAT content I'm entering and in WHICH application/field, not the mechanical act of pressing keys.

10. SCREEN CONTENT DETAIL. When I look at a computer or phone screen, describe WHAT is actually visible:
   - What app/website is open? What page/tab am I on?
   - What specific content is displayed? (article titles, code, chat messages, video titles, menu items)
   - For pop-up modals, dropdowns, autocomplete lists: enumerate the visible options/items.
   - GOOD: "I see a VS Code command palette showing options: 'Python: Select Interpreter', 'Python: Run File in Terminal', 'Format Document'."
   - BAD: "I examine the options listed in the pop-up modal." (WHAT options? List them!)

11. GAZE TRACKING. The green circle on each frame shows my exact gaze position. Use this to determine what I am actually looking at vs. what is merely visible in the periphery.

## Output Format
Return a JSON object:
{
  "scene_summary": "First-person one-sentence overview",
  "segments": [
    {
      "time_range": "HH:MM:SS-HH:MM:SS",
      "action": "I grab a coke from the fridge.",
      "action_brief": "I grab a coke.",
      "objects": ["specific objects I interact with or look at"],
      "environment": "FIRST segment: full establishing world state per Rule 4 (physical space, every visible object with state and position, complete screen content of every display, world text/signage, media playing, ambient audio). LATER segments: ONLY what changed since the previous segment, ignoring changes caused merely by head movement; 'No change.' is valid.",
      "text_visible": ["meaningful readable text near gaze point -- skip trivial keyboard/packaging labels"],
      "speech": "exact quote of what I or others say in this segment, or null",
      "details": "fine-grained context: which hand, which direction, micro-actions"
    }
  ],
  "activity_chain": "comma-separated first-person atomic actions"
}

## VERBATIM SPEECH -- hard requirement
Whenever I speak, the `action` field MUST carry my words **verbatim, character for character**,
inside quotes. Not paraphrased, not summarised, not truncated, not translated, not cleaned up.
- Keep the original language exactly as spoken (Chinese stays Chinese, English stays English,
  code-switching stays mixed).
- Keep filler, repetition, stutters and self-corrections as spoken ("Just help me enter a default
  default value first" keeps both default).
- Never replace any part of an utterance with "..." or "and so on" or a description of what I said.
  "I explain the requirements" is WRONG. `I say '<exact words>'` is the only acceptable form.
- The same verbatim quote also goes in the `speech` field. Both fields carry it in full.
- This applies to `action_brief` too: if the segment is a speech act, the quote survives
  compression intact -- drop the surrounding scaffolding, never the words themselves.

## action_brief -- Condensed Action
Every segment MUST also carry an `action_brief`: `action` with the dead weight removed.

THE RULE -- one verb, but keep everything that carries information:
- Exactly ONE main verb. When `action` chains verbs with "and"/"while"/"then", keep the single
  most informative one and drop the others. ("bring ... and set" -> "place"; "type ... and send"
  -> "send"; "rest my hand while watching" -> "monitor".)
- KEEP every information-bearing element, however long that makes it:
  * quoted speech or message text -- verbatim, character for character, never paraphrased,
    truncated or elided; see the VERBATIM SPEECH rule above
  * who it is addressed to / who is speaking
  * the specific topic, title, or subject matter
  * app / site / brand names, and the specific object being acted on
- DROP only what carries no information:
  device names ("on my MacBook"), body parts and manner ("with my right hand", "using the
  on-screen keyboard"), posture and location filler ("while sitting at my desk"), screen
  scaffolding ("on the screen", "in the input box"), and any verb already implied by another.
- This is compression, never invention or summarisation. Do not replace a specific noun with a
  generic one -- "the Claude Code explanation about TanhTransformedDistribution" must NOT become
  "explanations". No fixed word budget: as short as possible, but not one bit of signal shorter.

CALIBRATION -- these four are the standard:
- action: "I read the Claude Code panel explanation about TanhTransformedDistribution and
  change-of-variables log probability calculation."
  action_brief: "I read the Claude Code explanation about TanhTransformedDistribution and
  change-of-variables log probability."
  <- keep the topic; it is the whole point of the segment. Only "panel" and the trailing
    "calculation" go.
- action: "I bring a clear glass bottle to the countertop water dispenser beside the sink and
  set it on the dispenser tray."
  action_brief: "I place a glass bottle."
  <- two verbs -> one; the dispenser/sink/tray are scenery, the bottle is the object.
- action: "I type 'AI feels so slow' into the WeChat chat with 'User X' using the on-screen keyboard and
  send it."
  action_brief: "I send 'AI feels so slow' to User X."
  <- keep BOTH the message text and the recipient; drop the keyboard and the app chrome.
- action: "I rest my hand on top of the water dispenser while watching the glass bottle fill
  with water."
  action_brief: "I monitor the water filling."
  <- the hand is incidental; the watching is the action.

BAD (over-compression -- loses signal):
- "I read Claude explanations."            <- topic destroyed
- "I send a WeChat message."               <- message text and recipient destroyed
- "I interact with the interface."         <- everything destroyed

## Segmentation Rules
- Each segment = one meaningful action unit. Merge continuous/repetitive actions into one segment.
- A new segment starts when: a NEW action begins, I shift attention significantly, a new UI element appears, or someone speaks.
- Speech segments are separate from action segments. Every utterance MUST appear with exact quote -- but omit noise/fragments.
- Environmental observation segments ("I see...", "I hear...") are separate from action segments.
- Media content (videos I watch, audio I hear) -> environment segments with specific content details.
- Typical: 5-12 segments for a 30-second clip. Don't over-segment static or repetitive scenes.
\end{lstlisting}

\subsection{LLM-judge rubric}
We use three prompt variants across campaigns: P3 (Qwen), P4 (Gemini and GPT, originally in Chinese), and P6 (Haiku). They share the same evaluation dimensions---action-level similarity in verb, object, and temporal alignment---but differ in example language and scoring anchors. P4 is printed below as a representative; the full set is archived in \texttt{prompts/judge\_campaigns\_original.json}.
\paragraph{P4: Gemini and GPT campaigns (translated).}
\begin{lstlisting}
You are an expert in behavioral semantic similarity. Determine whether Prediction and Ground Truth describe the same behavior.

Scoring (integer 0-9):
9 = Exact match: core verb, object, and intent all match; synonyms and near-equivalent expressions count as matches.
7-8 = Highly similar: correct action, with a slightly generalized object or different details.
4-6 = Partially related: prerequisite action or same location/object, but different or incomplete ultimate intent.
2-3 = Weakly related: only approximately shared location or scene.
0-1 = Unrelated: completely different behaviors.

Examples:
GT: "I pick up my phone"; Pred: "I take out my phone to check messages" -> 7 (same verb and object, with an added purpose).
GT: "I walk toward the kitchen"; Pred: "I move closer to the kitchen" -> 9 (perfect paraphrase).
GT: "I photograph the cake"; Pred: "I cut the cake into small pieces" -> 1 (completely different behaviors).

Output only one integer from 0 to 9, with no explanation.
\end{lstlisting}

\end{document}